\documentclass{article}
\usepackage{ijcai20}

\usepackage{times}
\usepackage{soul}
\usepackage{url}
\usepackage[hidelinks]{hyperref}
\usepackage[utf8]{inputenc}
\usepackage[small]{caption}
\usepackage{graphicx}
\usepackage{booktabs}
\usepackage[ruled,linesnumbered]{algorithm2e}
\SetAlCapNameFnt{\footnotesize}
\SetAlCapFnt{\footnotesize}
\renewcommand\footnotemark{}
\usepackage{multirow}
\usepackage{multicol}
\usepackage{makecell}
\usepackage{amsmath,amssymb,amsthm,amsfonts,mathrsfs,bm,multirow}

\DeclareMathOperator*{\minimize}{\text{minimize}}

\usepackage{xcolor}
\definecolor{xiaolong}{RGB}{181, 97, 2}

\title{BLK-REW: A Unified Block-based DNN Pruning Framework using Reweighted Regularization Method}

\author{
Xiaolong Ma\text{$^\dagger$}$^1$\and
Zhengang Li\text{$^\dagger$}$^1$\thanks{\hspace{-6.0mm}$^\dagger$These authors contributed equally.}\and
Yifan Gong$^{1}$\and
Tianyun Zhang$^{2}$\and
Wei Niu$^3$\and
Zheng Zhan$^{1}$\and \\
Pu Zhao$^{1}$\and
Jian Tang$^{4}$\and
Xue Lin$^{1}$\and
Bin Ren$^3$\and
Yanzhi Wang$^{1}$
\affiliations
$^1$Northeastern University, 
$^2$Syracuse University, 
$^3$College of William and Mary,
$^3$DiDi AI Labs
\emails
ma.xiaol@husky.neu.edu, yanz.wang@northeastern.edu
}

\begin{document}

\maketitle

\begin{abstract}
Accelerating DNN execution on various resource-limited computing platforms has been a long-standing problem. Prior works utilize $l_1$-based group lasso or dynamic regularization such as ADMM to perform structured pruning on DNN models to leverage the parallel computing architectures. However, both of the pruning dimensions and pruning methods lack universality, which leads to degraded performance and limited applicability. To solve the problem, we propose a new block-based pruning framework that comprises a general and flexible structured pruning dimension as well as a powerful and efficient reweighted regularization method. Our framework is universal, which can be applied to both CNNs and RNNs, implying complete support for the two major kinds of computation-intensive layers (i.e., CONV and FC layers). To complete all aspects of the pruning-for-acceleration task, we also integrate compiler-based code optimization into our framework that can perform DNN inference in a real-time manner. To the best of our knowledge, it is the first time that the weight pruning framework achieves universal coverage for both CNNs and RNNs with real-time mobile acceleration and no accuracy compromise.
\end{abstract}

\section{Introduction}

% Due to their high accuracy and scalability, Deep Neural Networks (DNNs) such as Convolutional Neural Networks (CNN\textcolor{cyan}{s}) and Recurrent Neural Networks (RNN\textcolor{cyan}{s}) have become the key component of artificial intelligence \textcolor{cyan}{(AI)} systems. DNN models are deployed in various AI systems for inference and recognition tasks.
% However, the high accuracy of DNNs typically comes with high computation cost and storage requirement. Accelerating DNN inference is therefore of great importance to many AI applications, especially those with critical time constraints, such as self-driving cars~\cite{makantasis2015deep} and real-time translation~\cite{gehring2016convolutional}.

% Deep Neural Networks (DNNs) such as Convolutional Neural Networks (CNNs) and Recurrent Neural Networks (RNNs) have been extensively adopted in various artificial intelligence (AI) systems due to their high accuracy and scalability. However, the high accuracy typically comes with high computation cost. Accelerating the DNN inference is thus crucial for many AI applications, especially those with critical time constraints, such as self-driving cars~\cite{makantasis2015deep} and real-time translation~\cite{gehring2016convolutional}.

Deep Neural Networks (DNNs) such as Convolutional Neural Networks (CNNs) and Recurrent Neural Networks (RNNs) have been extensively adopted in various artificial intelligence (AI) systems. However, accelerating the computational intensive DNN inference is very challenging for many AI applications, especially those with critical time constraints, such as self-driving cars~\cite{nugraha2017towards} and real-time translation~\cite{gehring2016convolutional}.

Pruning has gained its popularity due to the effectiveness in reducing model size and computation cost. 
In order to remove redundant weights while maintaining accuracy, many studies have been proposed regarding both pruning dimension (DNN structure level) and pruning method (algorithm level). 
According to the structure of pruned models, there are mainly two DNN pruning approaches: {\em non-structured pruning} and {\em structured pruning}. 
However, non-structured pruning has been proven by many recent studies~\cite{wen2016learning,he2017channel} that it is not compatible with the parallelism in hardware accelerations due to the imbalanced computation and significant overhead. 
Structured pruning has been proposed to conquer the challenge. 
% By generating different kind of structured sparsity, such as filter-wise sparsity, shape-wise sparsity and channel-wise sparsity, 
A structured pruned model maintains the regularity of the weight matrix, which eliminates the overhead and facilitates on-device acceleration. However, the aggressive pruning strategy causes severe information loss, making accuracy degradation non-negligible. Achieving both high accuracy and fast inference with DNN pruning is an ideal but very challenging goal.

Efforts have been made to achieve this goal. At algorithm level, many pruning techniques have been proposed to find the uncritical weights. For non-structured pruning, prior works leverage a magnitude-based pruning method that prunes weights with small magnitudes or use $l_1$ regularization to explore sparsity in DNN models. For structured pruning, the static $l_1$-based group lasso regularization is used to find the regular sparse pattern in DNN models. 
However, the above approaches fail to find a satisfactory solution for the pruning problem due to the heuristic nature. ADMM~\cite{boyd2011distributed} algorithm emerges to mitigate the challenges. 
With a significant improvement in the solution quality, ADMM pruning supersedes (almost) every pruning framework and becomes the state-of-the-art method. Nevertheless, ADMM still suffers from sub-optimal solution quality and long convergence time, especially for the long-standing problem of finding structured sparsity solution for the Fully Connected (FC) layer. This will certainly limit the usage of ADMM solutions on many CNNs and almost all RNNs since they are majorly composed of FC layers.

In this paper, we present a unified pruning framework -- block-based structured pruning with reweighted regularization (BLK-REW). Our efforts focus on two aspects: pruning dimension and pruning method. 

\noindent {\bf Aspect 1:} From the pruning dimension aspect, we propose block-based structured pruning (BLK pruning) which divides DNN layers into multiple blocks and applies structured pruning independently to each block. Our design takes a unique perspective on structured pruning, which greatly enlarges the design space by introducing a higher degree of flexibility with a changeable block shape. More importantly, the proposed BLK pruning is applicable to both CNNs and RNNs without obvious accuracy degradation, which outperforms the existing pruning dimensions. It achieves similar or even higher accuracy compared with non-structured pruning, and preserves the hardware compatibility advantage of structured pruning, with the compiler-based code optimization embedded in our pruning-acceleration framework.

% \textcolor{cyan}{from pruning dimension aspect, we propose block-based structured pruning (BLK pruning) which divides each DNN layer into multiple blocks and applies structured pruning independently to each block. Our design is more flexible and provides finer granularity, thereby it can attain higher accuracy, compared with conventional structured pruning dimensions. More importantly, the proposed BLK pruning is applicable to both CNNs and RNNs without obvious accuracy degradation. We will show that BLK pruning outperforms the existing pruning dimensions. On the one hand, it achieves similar or even higher accuracy compared with non-structured pruning. On the other hand, it preserves the advantage of structured pruning, i.e., the compatibility with hardware inference acceleration, with the compiler-based accelerator embedded in our pruning-acceleration framework.}

\noindent {\bf Aspect 2:} From the pruning method aspect, we propose to use reweighted (REW) group lasso regularization method to generate structured sparsity. By introducing a reweighted term into regularization, our method can perform group regularization at a more precise location in DNN with an appropriate degree. 
Compared with the traditional $l_1$-based group lasso and the recently developed ADMM regularization method, 
REW method acquires a significant improvement in the regularization effect (i.e., facilitating better pruning results) with a desirable short convergence time (i.e., efficient training process), 
which makes it a favorable approach that naturally fit for the DNN pruning problems. 
% of finding sparsity in DNN models which naturally fit for the pruning problem, 
% thereby achieving better results.

We show the performance improvements of BLK-REW framework in three ways. \emph{First}, the proposed REW method can efficiently find uncritical weights. Compared with other methods, REW achieves better weight regularization effect using significantly shorter training time. 
\emph{Second}, the proposed BLK pruning dimension is more general and achieves extremely high compression rates in both CNN and RNN. 
% For large-scale DNN structure such as VGG-16, it achieves \textcolor{red}{50$\times$} weight reduction which outperforms almost every prior art. 
\emph{Third}, the proposed BLK-REW pruning naturally fits for the compiler optimization. Our designed compiler-aided acceleration framework achieves real-time inference on the resource-limited mobile devices.

\vspace{-2mm}
\section{Background and Motivation}
% As the most straightforward and efficient neural network compression technique, weight pruning removes redundant or less important weights to reduce the storage and computation costs. According to the structure of pruned models, there are mainly two DNN pruning approaches: {\em non-structured pruning} and {\em structured pruning}. 
% However, non-structured pruning has been proven by many recent studies \textcolor{red}{[REFs]} that it is not compatible with the parallelism in hardware accelerations due to the imbalanced weight distribution and associated huge computation overhead. 
% Thus, in this paper, our study focuses on structured weight pruning and corresponding regularization-based pruning methods. 

\subsection{Structured Pruning Dimension}
\begin{figure}[t]
    \centering
    \includegraphics[width=0.45 \textwidth]{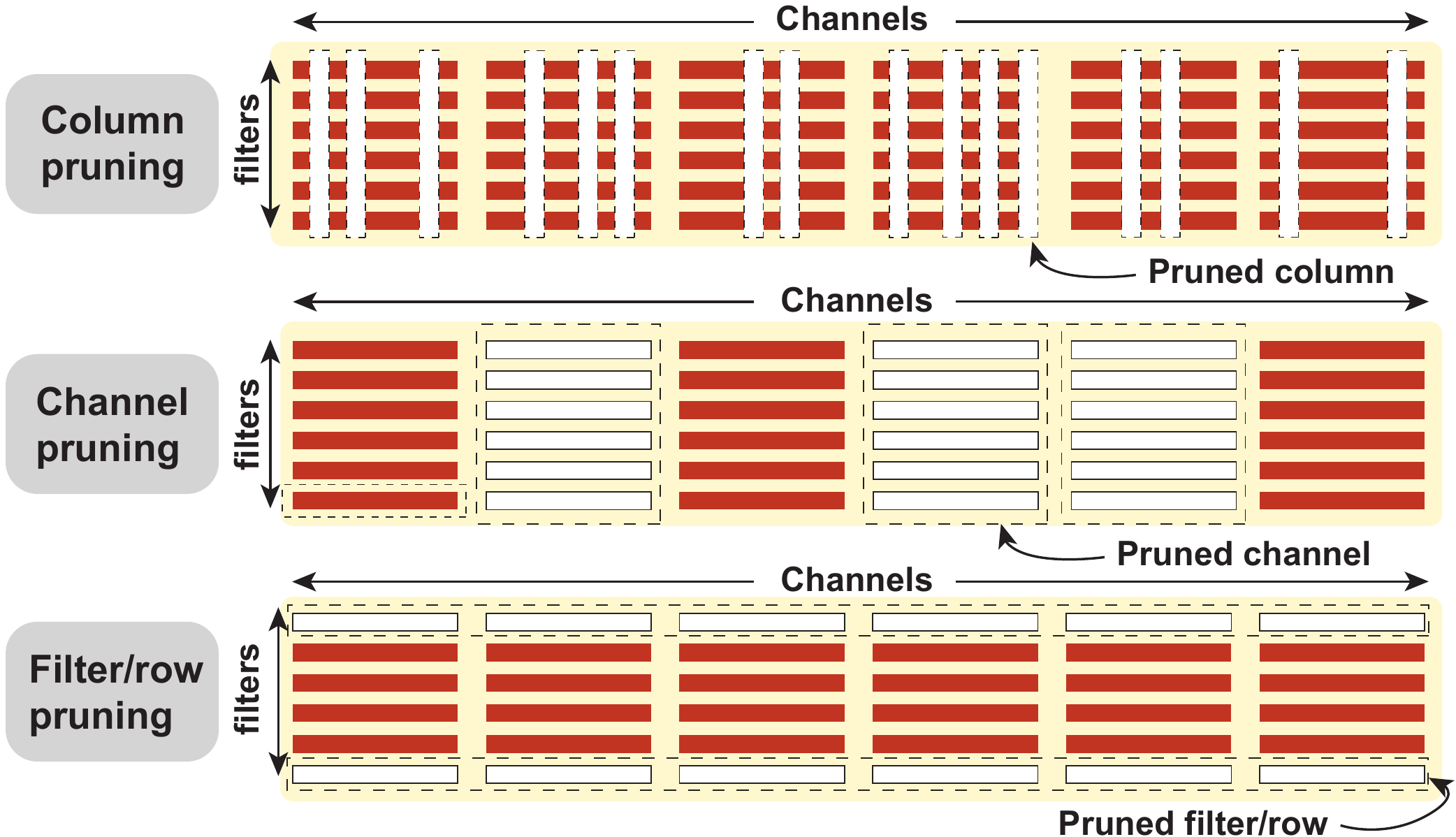}
    %\vspace{-1mm}
    \caption{Different types of structured pruning.}
    \label{fig:structured}
    \vspace{-3mm}
\end{figure}

Recent works~\cite{wen2016learning,he2017channel} considered to incorporate regularity (i.e., filter pruning, channel pruning, etc.) in weight pruning, which generates regular and smaller weight matrices for faster executions on CPUs/GPUs. 
For convolution computations, weight matrices are usually transformed into general matrix multiplication (GEMM) form as Figure~\ref{fig:structured} illustrates. Accordingly, filter pruning can also be termed as row pruning since it corresponds to removing one row of the weight matrix, and channel pruning corresponds to reducing multiple consecutive columns (column pruning). Current structured pruning approaches suffer from notable accuracy loss when the compression rate is high because the entire information of the pruned filter(s)/channel(s) is lost. As a result, it usually has limited compression rates and low accuracy, as well as limited applicability as most works focus on CONV layers only. For FC layers (applied partially in CNN and majorly in RNN), structured pruning is applicable but not desirable due to the same reason above. The drawback is obvious, especially for time-based RNNs since one pruned row/column in an RNN will not be utilized for all timestamps, causing server accuracy degradation.

\subsection{Regularization-based Pruning Methods}
Finding structured sparsity in a DNN model is intrinsically solving an $l_0$ optimization problem with structured constraints. Two mainstream methods have been proposed to solve this problem. One incorporates a static regularization term into DNN training, and the other one uses a dynamically updated regularization term during DNN training. 

{\bf Static regularization} is firstly utilized in solving non-structured pruning problems by incorporating $l_1$ regularization into DNN training. By extending $l_1$ regularization into group lasso~\cite{yuan2006model,wen2016learning,he2017channel} form, structured pruning on DNN models can also be achieved. 
% usually achieved by using regularization methods, such as $l_1$ regularization and group lasso~\cite{yuan2006model,wen2016learning,he2017channel} regularization. $l_1$ regularization is first adopted in solving non-structured pruning problems. By extending it into group lasso form, structured pruning on DNN models can also be achieved. 
With specified regularization dimensions (groups), it can perform different types of structured pruning (i.e., filter pruning, channel pruning and the combination of them). However, this method yields limited compression rates and non-negligible accuracy degradation due to the intrinsically heuristic and non-optimized approach. 

{\bf Dynamic regularization} method such as ADMM pruning~\cite{zhang2018systematic,ren2019ADMMNN} usually reforms pruning problems into optimization problems with dynamically updated regularization terms bounded by the designated constraint (i.e., pruning with specific dimensions or with any desired weight matrix shapes) sets.
During training, ADMM can separately and iteratively solve the pruning problem. 
Although this method is revolutionary in its functionality and outperforms the former ones in terms of pruning rate/accuracy, a satisfactory solution cannot always be guaranteed for the non-convex (i.e., DNN loss function) problem, not to mention that this method suffers from a time-consuming training process.

\subsection{Motivation}\label{sec:motivation}
% Our motivation originates from the shortcomings of the aforementioned two aspects. 

{\bf From the pruning dimension aspect}, 
the current structured pruning dimensions suffer from major information loss. The accuracy drop is especially significant in RNN pruning. 
The motivation of our study is to seek an approach to maintain the regularity in the pruned model (for facilitating hardware acceleration), while restoring the flexibility of the spatial distribution of the weights (to re-gain high accuracy). In our proposed BLK pruning which is applicable to both CNNs and RNNs, we take a unique step towards this goal by introducing a new pruning perspective, and avoid the pitfall of making this approach ``a mere trade-off" between model accuracy and regularity. 
We also take a further step of compiler optimization to establish the connection between the general, BLK sparsity and the on-device speedups. 
Integrating all merits into one design, the accuracy can be similar or even surpass the non-structured pruning, and the on-device acceleration performance can be close to structured pruning.

{\bf From the regularization aspect}, 
we emphasize that both current static and dynamic regularization methods are limited by their intrinsic shortcomings. 
For static regularization, the $l_1$ or group lasso regularization applied on the loss function penalizes all weights in its dimension scope through the entire network, which means some important weights are penalized to near-zero values, thereby resulting in highly impaired solutions. On the other hand, the dynamic regularization method reforms pruning problem as an optimization problem with hard constraints on $l_0$ norm, and then use ADMM to solve it. 
However, this method suffers from long convergence time due to the strong non-convexity of $l_0$ norm, especially with structured hard constraints. Using ADMM in the training process also inevitably generates extremely small weights that are difficult to remove, not to mention the hard constraints cause a large amount of hyper-parameters that need to be tuned manually for each layer, which is very inefficient. It is imperative to find an effective method to solve the $l_0$ optimization problem with self-adaptive regularization and soft constraints.

\section{Unified and Flexible Framework of DNN Pruning - Acceleration}

In this section, we propose a unified framework of DNN weight pruning, supporting (i) the flexible, \emph{block-based structured (BLK) pruning} that applies to both CNN and RNN architectures, and (ii) highly effective weight pruning algorithm with \emph{reweighted (REW) method}. 
Our framework also includes a general method to accelerate DNN execution by utilizing compiler-based code optimization, achieving holistic supports for the DNN pruning-acceleration studies.

\subsection{Block-based Structured Pruning -- A Unique Perspective on Structured Weight Pruning} 
Conventional, structured pruning treats the DNN weight matrix in each layer as a whole, and selects to prune a whole row or column of the entire weight matrix. However, the accuracy performance is hindered by this limited, inflexible view of structured pruning. 

\begin{figure}[t]
    \centering
    \includegraphics[width=0.48 \textwidth]{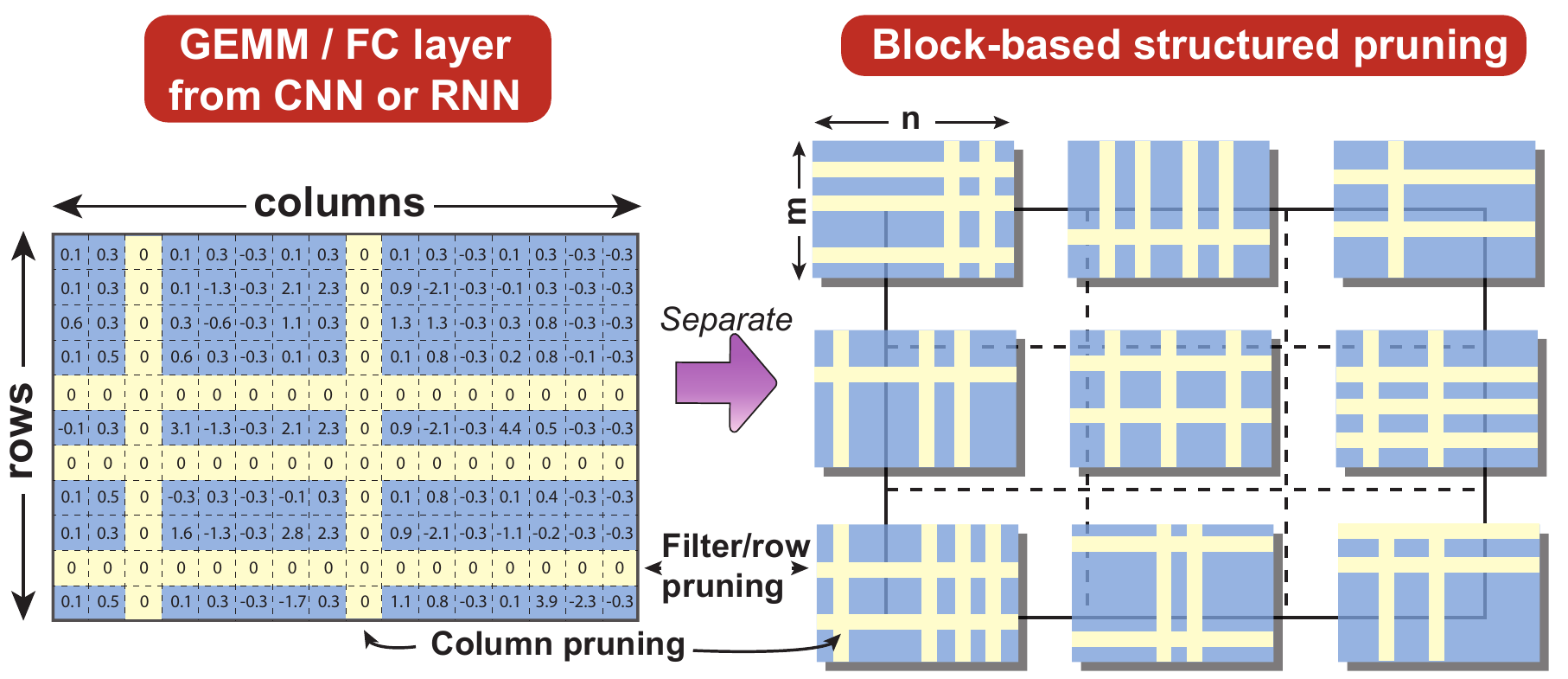}
    %\vspace{-1mm}
    \caption{Proposed flexible, block-based structured pruning.}
    \label{fig:blk}
    \vspace{-5mm}
\end{figure}

In our perspective, we consider the weight matrix in each layer (e.g., GEMM or FC that represent different types of layer-wise computation) to be composed of multiple weight blocks with the same size $m\times n$ as Figure~\ref{fig:blk} shows. 
We apply independent row and column pruning on each block, with potentially different pruning rates (number of pruned rows/columns) in each block, to ensure high flexibility. The remaining weights in each block still form a full matrix with a smaller size. 
Within our perspectives, the aforementioned non-structured pruning and the state-of-the-art structured pruning are two extremes in our design with 
the block size $1\times 1$ (i.e., non-structured pruning) and the size of the whole matrix (i.e., structured pruning). 

\begin{figure}[h!]
    \centering
    \includegraphics[width=0.4 \textwidth]{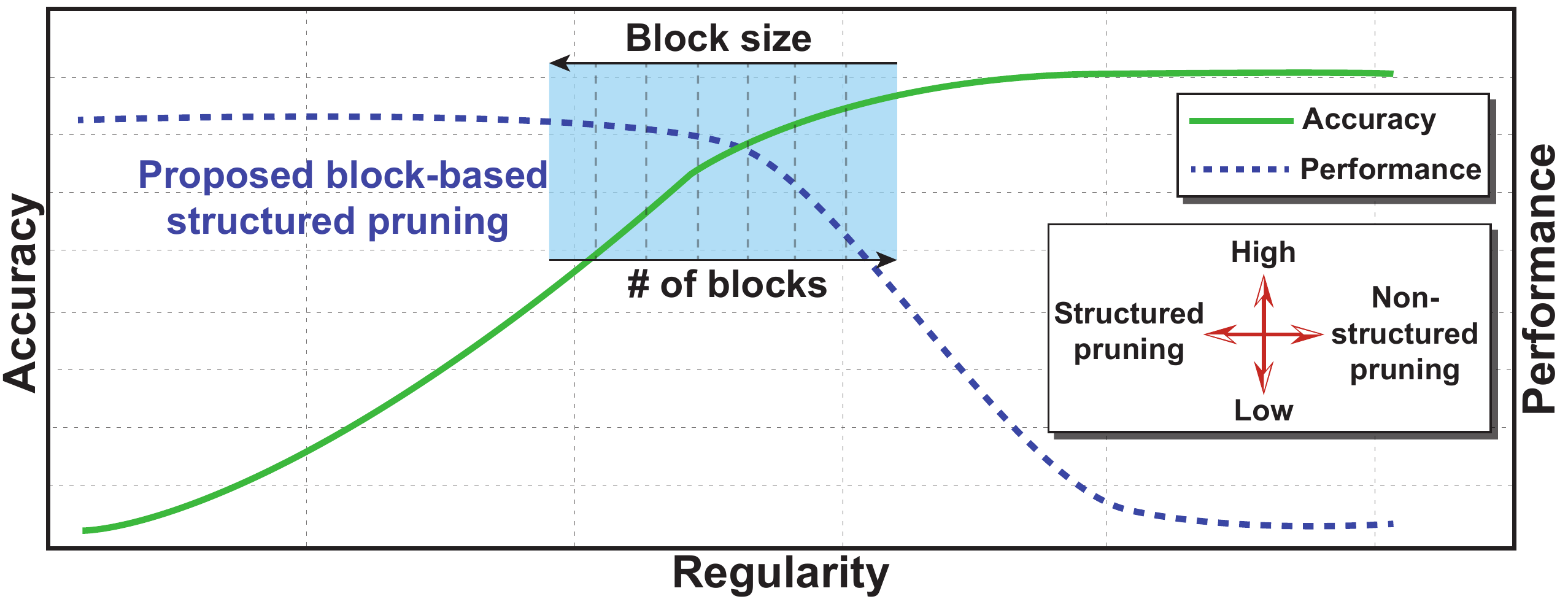}
    %\vspace{-1mm}
    \caption{An illustrative demonstration of the the regularity and accuracy of the proposed block-based structured pruning.}
    \label{fig:illustratuve}
    \vspace{-4mm}
\end{figure}

We will discuss in our experiment results that the BLK pruning is not just ``a mere trade-off" as Figure~\ref{fig:illustratuve} shows. 
The reason is that pruning is processed within each block independently, thereby part of the weights with important information in each filter/channel is preserved, implying high flexibility. In the meantime, the remaining weights still maintain a certain degree of regularity. 
It is beneficial to both DNN accuracy and inference acceleration. 
The high flexibility and regularity enabled by our approach reveal a huge design space that potentially facilitates versatile front-end systems. 

% \textcolor{red}{TODO: illustrative Figure for performance and accuracy.}

% \textcolor{blue}{Data and figure to show that? Also you need to show acceleration results in the experiemental results.}

\subsection{Effective Regularization-based Pruning Algorithm with Reweighted Method}
For an $N$-layer DNN of interest, let $\mathbf W$ denote the collection of weights for $i$-th layer, i.e., $\mathbf W = \{ \mathbf W_i \}_{i=1}^N$. According to our design of the flexible, block-based sparsity, we propose the following $l_0$ constraints on the pruning of $\mathbf W_i$.

{\tt Constraints:} Each $\mathbf W_i$ will be uniformly divided into $K$ blocks with the size of $m\times n$ in each of the GEMM or FC matrix, namely, $\mathbf W_i = [\mathbf W_{i1},  \mathbf W_{i2},...,\mathbf W_{iK}]$, where $\mathbf W_{ij} \in \mathbb R^{m \times n}$. Let $[\mathbf W_{ij}]_{p,:}$ and $[\mathbf W_{ij}]_{:,q}$ denote the $p$-th row and the $q$-th column of $\mathbf W_{ij}$, respectively. 

Towards training of the DNN, we minimize the loss function of the network to increase accuracy. 
In order to achieve structured sparsity, the common method is to add group lasso regularization~\cite{yuan2006model} to the loss function. In fact, achieving block-based row and column sparsity is also a special group lasso problem. 
% In our case, we propose to use reweighted method on group lasso regularization. 
Let $f({\mathbf W})$ denote the training loss. 
The classic optimization with group lasso regularization on the block-based sparsity can be formulated as
\begin{align}
\begin{array}{cl}
\displaystyle \minimize_{\mathbf W} f({\mathbf W}) +  \lambda  \sum_{i=1}^N  \sum_{j=1}^K 
 \| [\mathbf W_{ij}]  \|_{g} 
\end{array}
\label{eq: grouplasso}
\end{align}
where $\lambda$ is the penalty parameter to adjust the relative importance of accuracy and sparsity degree, and $\|\cdot \|_{g}$ denotes group lasso computation. 
It is difficult to find high quality solution using this fixed regularization method (please refer to the explanation in Section~\ref{sec:motivation}). Instead, an effective dynamic regularization method dealing with such soft constraints is in need. To achieve this goal, we propose to use reweighted method~\cite{candes2008enhancing} to solve group lasso regularization, thereby eliminating the previous shortcoming of applying the same penalty on important and less significant weights. 
We formulate the following two optimization problems for block-based row pruning and column pruning. 

{\bf For block-based row pruning}, we solve
\begin{align}
\begin{array}{cl}
\displaystyle \minimize_{\mathbf W} f({\mathbf W}) +  \lambda  \sum_{i=1}^N  \sum_{j=1}^K \bigg(
{\mathcal P}_{i}^{(t)} \circ \| [\mathbf W_{ij}]_{p,:}  \|_2  \bigg ) 
\end{array}
\label{eq: rew_row}
\end{align}
where $\circ$ denotes element-wise multiplication, $\| \cdot \|_2$ denotes the Frobenius norm and ${\mathcal P}_{i}^{(t)}$ is the collection of penalty weights\footnote{${\mathcal P}$ is initialized by the original weights in the pre-trained model.}, which is updated in every iteration $t$ to help increase the degree of sparsity beyond group lasso regularization. In each iteration, the solution of ${{\mathbf W}}_{i}$ is given by ${{\mathbf W}}_{i}^{(t)}$ and we update ${\mathcal P}_{i}$ by setting 
\[
{\mathcal P}_{i}^{(t+1)} = \frac{1}{\| [({\mathbf W}_{ij})]_{p,:}^{t} \|_2^2+{\epsilon}}
\]
where $\epsilon$ is a parameter with small value to prevent the division by zero denominator. 

{\bf For block-based column pruning}, we solve
\begin{align}
\begin{array}{cl}
\displaystyle \minimize_{\mathbf W} f({\mathbf W}) +  \lambda  \sum_{i=1}^N  \sum_{j=1}^K \bigg(
{\mathcal P}_{i}^{(t)} \circ \| [\mathbf W_{ij}]_{:,q}  \|_2  \bigg ) 
\end{array}
\label{eq: rew_col}
\end{align}
and update ${\mathcal P}_{i}$ by 
\[
{\mathcal P}_{i}^{(t+1)} = \frac{1}{\| [({\mathbf W}_{ij})]_{:,q}^{t} \|_2^2+{\epsilon}}
\]
Please note that \eqref{eq: rew_row} and \eqref{eq: rew_col} can be solved separately or simultaneously using the standard solver.

\begin{algorithm}[htp]
\footnotesize
\caption{Reweighted regularization for block-based structured pruning}
\label{alg:rew}
\textbf{Initialization:} Pretrained DNN model with initialized $\mathcal P_{i}$; Set $t\gets 1$ and total iteration number $T$; Pre-define block size $m$ and $n$ \\ 
\KwResult{Block-based structured pruned model;}
% Pre-define the block size $m$ and $n$; \\
% Calculate $K$ for each layer based on the value of $m$ and $n$. \\
Each layer $K\gets$ (the size of ${\mathbf W}_{i}$)/($m\times n$); \\
\While{$t \le T$}{
Solve \eqref{eq: rew_row} and/or \eqref{eq: rew_col} using standard solver in SGD \;
Update ${\mathcal P}_{i}^{(t+1)}$ using the solution of ${\mathbf W}_{i}^{(t)}$
}
Remove the group of weights close to zero and retrain the rest of non-zero weights to refine accuracy.
\end{algorithm}
Algorithm~\ref{alg:rew} describes the general steps that are used in the proposed REW method. We first initialize $\mathcal P_{i}$ using the pretrained model, and pre-define the block size for the pruned model. During DNN training, we incorporate the reweighted group lasso regularization in \eqref{eq: rew_row} and \eqref{eq: rew_col}, and update the penalty parameter ${\mathcal P}_{i}$ iteratively. By updating the penalty, we ``\emph{reweight}" the regularization term(s) that is (are) bounded in the optimization problems. 
After reweighted steps, we remove the weights (or group of weights) which are close to zeros and refine the DNN using the non-zero weights.

\textbf{Reweighted regularization analysis:} 
Consider that two weights $w_i$ and $w_j$ ($w_i<w_j$) are penalized by certain regularization. The larger $w_j$ is inevitably being penalized more heavily than the smaller $w_i$. Although it is easier for $w_i$ to become zero, the fact that $w_j$ is penalized still violates the original intention of weight pruning, which is to remove the ``uncritical" weights. Larger weights typically serve a critical role in generating stronger activation for a more confident decision. In the REW method, $w_j$ remains un-penalized or even being rewarded while $w_i$'s penalty is amplified. 
Interestingly, our experimental results in Section~\ref{sec:critical_dist} show that the importance of a (group of) weight is also related to its location, and the REW method can effectively separate those locations. We claim that this characteristic is attributed to the systematic and iterative manner of the REW method.

% \textcolor{red}{Speed or number of epochs? Faster than ADMM? Reduction in the number of hyperparameters?}

\textbf{Reweighted training:} Compared with ADMM training which also uses an iteratively updating scheme for the regularization term, the proposed REW method uses fewer training epochs for the loss to converge. For example, when pruning VGG-16 on CIFAR-10, the ADMM method usually requires 1,000 - 1,200 epochs to converge when the compression rate is around 20$\times$. Additionally, the retraining step also requires the same amount of epochs to restore accuracy. In the proposed reweighted training, we only need 150 - 200 epochs for reweighted step and 200 epochs for retraining. In the meantime, ADMM requires setting pruning ratio and other hyper-parameters (e.g., layer-wise penalty) for each layer manually, while the proposed REW method only requires one penalty parameter for all layers. Also, the soft constraints in REW method determine pruning ratio for the whole network automatically, which eliminates a lot of parameters that need to be set empirically.

\textbf{Multiple objective functions:} The original objective function in the proposed REW method is targeting at DNN weight reduction. However, our objective function can also be formulated for operation (FLOPS) reduction, storage reduction, etc., and solved using the same REW method. 
Due to space limits, those formulations will not be discussed.

\begin{figure}[t]
    \centering
    \includegraphics[width=0.48 \textwidth]{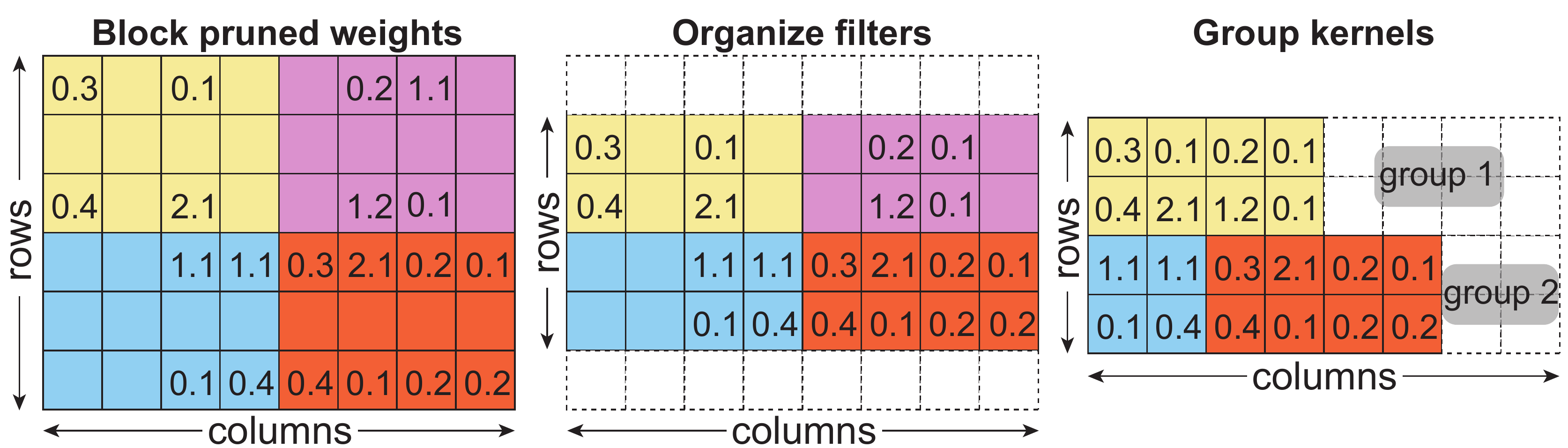}
    %\vspace{-1mm}
    \caption{Compact weights by matrix reorder.}
    \label{fig:fkr}
    \vspace{-4mm}
\end{figure}

\begin{figure*}[t]
    \centering
    \includegraphics[width=0.99 \textwidth]{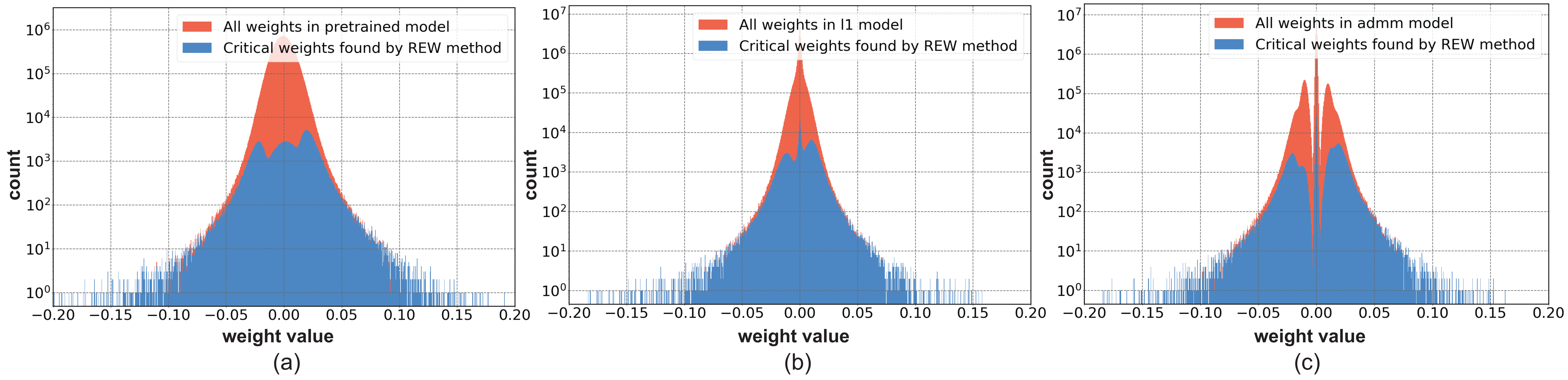}
    % \vspace{-1mm}
    \caption{Critical weights distribution (logarithmic scale) found by reweighted regularization method in the first FC layer of a VGG-16 model. The comparison includes (a) a pretrained model, (b) an $l_1$-based group lasso regularized model and (c) an ADMM regularized model.}
    \label{fig:critical}
    \vspace{-3mm}
\end{figure*}

\subsection{Compiler-aided Mobile Acceleration Framework for Block-based DNN Sparsity}
In order to fully leverage the block-based sparsity, we design a compiler-aided acceleration framework to deploy DNN models on the computing platform. 
We adopt code generation to convert a DNN model into computational graph which is embodied by static C++ (for CPU execution) or OpenCL (for GPU execution) code, 
and with the optimization techniques to guarantee end-to-end execution efficiency. 
This work use mobile devices as the computing platform. However, the concept and principle of using compiler to execute DNN is universal and can be utilized in (almost) every computing device.  

The compiler optimization aims to address the following performance challenges in pruned DNN executions: {\em thread divergence} and {\em load imbalance} caused by the well-known challenges of the sparse matrix multiplications. To mitigate the challenge, we propose the \emph{matrix reorder} technique.

{\bf Matrix reorder:} At first glance, the block-based sparsity has a disordered weight distribution, which incurs significant thread divergence and load imbalance if rows are processed by different threads. Figure~\ref{fig:fkr} illustrates our proposed matrix reorder technique. 
As the remaining weights that appear in certain rows and columns in each block have a certain degree of regularity, 
we first reorder the rows (e.g., filters in CNN) by arranging the ones with the same or similar patterns together. Next, we compact the weights in the column direction (e.g., kernels in CNN). At last, the rows with the same or similar computations are grouped together. 
As a result, each group is processed by all threads in parallel, and each thread is in charge of multiple consecutive rows. Thus, the computation divergence among these threads is significantly reduced. 
On the other hand, since the weight distribution pattern in each block is regular and \emph{known} after grouping, the input matrix that corresponds to each weight group will be loaded only once. The load imbalance can be relieved thanks to the register level loading operation reduction.

\section{Experimental Results}
{\bf Methodology:} In our experiment, the proposed BLK-REW pruning framework is utilized on two different machine learning tasks -- image classification and natural language processing (NLP). 
In image classification tasks, our experiments are based on four widely used CNN structures, VGG-16~\shortcite{simonyan2014very}, ResNet-18/50~\shortcite{he2016deep} and MobileNet-V2~\shortcite{howard2017mobilenets} on CIFAR-10 and ImageNet datasets; and for NLP task, we test our proposed pruning framework on GRU with TIMIT dataset. 
We train the networks on an eight NVIDIA Titan RTX GPUs server using PyTorch~\shortcite{paszke2019pytorch}. 

In order to show the acceleration of block-based sparsity on mobile devices, we compare it with three state-of-the-art DNN acceleration frameworks, TensorFlow-lite~(\citeauthor{TensorFlow-Lite}), TVM~\cite{chen2018tvm}, and MNN~(\citeauthor{Ali-MNN}). 
Our evaluations are conducted on a Samsung Galaxy S10 phone with the latest Qualcomm Snapdragon 855 that consists of a Qualcomm Kryo 485 Octa-core CPU and a Qualcomm Adreno 640 GPU.

\begin{figure*}[t]
    \centering
    \includegraphics[width=0.955 \textwidth]{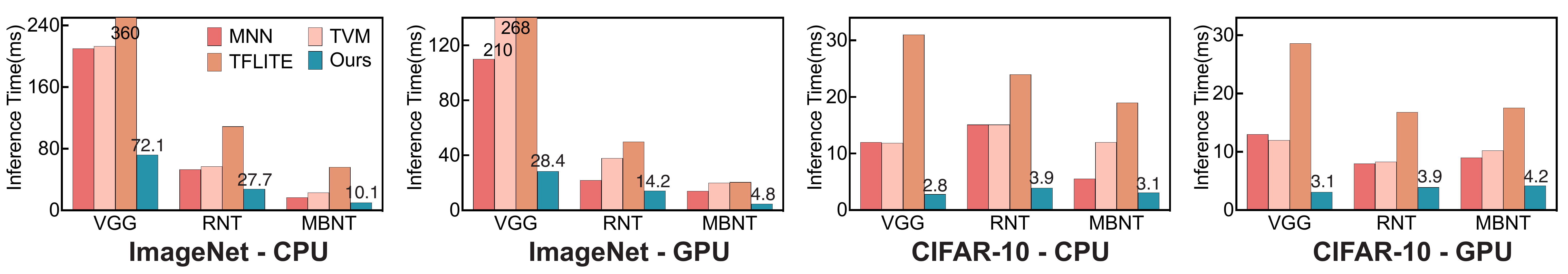}
    %\vspace{-1mm}
    \caption{Mobile CPU/GPU inference time ($ms$) on different network structures inferring CIFAR-10 and ImageNet images.}
    \label{fig:speed}
    \vspace{-3mm}
\end{figure*}

\subsection{Critical Weights Analysis on Different Regularization Methods}\label{sec:critical_dist}

We state that the proposed REW method can achieve better pruning result. The reason is that our method can effectively separate the uncritical weights from critical ones. 
We use VGG-16 on ImageNet to generate a sparse model based on the proposed reweighted regularization method, and compare it with $l_1$-based regularization as well as ADMM regularization. To ensure absolute fairness, all the models in the comparison use the same pruning dimension and compression rate. In this case, we use one block (i.e., prune entire columns and rows) in each layer for all methods. 

Figure~\ref{fig:critical} illustrates the difference of critical weights distribution between REW method and others. We first find the non-zero value \emph{positions} in the sparse model generated by our REW method. Through using those \emph{positions}, we find the corresponding weights and their distribution in (i) a pretrained model, (ii) an $l_1$-based group lasso regularized model and (iii) an ADMM regularized model. The critical weight distribution is shown in Figure~\ref{fig:critical}, with the orange color denoting original weights distribution and the blue color indicating the ``critical" weights found and preserved by our method. According to the figure, we have the following analyses: 

\noindent {\bf \emph{(a).}} In a pretrained DNN model, some weights with small magnitude are critical to maintain accuracy. Therefore, some pruning works that only prune small weights are very subjective and hard to achieve good results.

\noindent {\bf \emph{(b).}} In an $l_1$-based group lasso or ADMM regularized model, part of the weighs are penalized to zero or near-zero values, and then those close-to-zero values are pruned and the rest non-zero values are retrained to restore accuracy. However, the REW method considers some weights that have been penalized are critical, thus should not be pruned. 

We conclude that REW method separates critical weights in a very different way, in which the importance of weight(s) is not only based on its value, but also associated with its position. 
To prove and reinforce our conclusion, we need to show a strong accuracy improvement of the REW method compared with others, which is reported in the following section.

\subsection{Accuracy Analysis on Overall Model Compression Results}
In our previous analyses, we stress that reweighted regularization can effectively separate critical weights, thus achieving better pruning solutions. In this part, we demonstrate the overall compression results to support our conclusion. Specifically, we prune the entire rows and columns (i.e. using one block for each layer) with REW method to compare with other methods (e.g., lasso, ADMM and other heuristics). Beyond one block structured pruning, we also divide weights into several blocks to show BLK-REW pruning results. 

Table~\ref{tab:cifar} and Table~\ref{tab:imagenet} show our pruning results using different CNN structures with CIFAR-10 and ImageNet datasets. Table~\ref{tab:timit} shows RNN pruning results using GRU with TIMIT dataset. Overall, when we prune entire rows and columns using the proposed REW method, the compression results consistently outperform the baseline methods. By using the BLK-REW framework, we unprecedentedly achieve better compression results for both CNNs and RNNs, leading to lightweight model size and computation. 

\begin{table}[t]
% \scriptsize
    \centering
    \caption{BLK-REW pruning results on CIFAR-10 using VGG-16 and ResNet-18 and MobileNet-V2 (MBNT).}
    \resizebox{0.48 \textwidth}{!}{
        \begin{tabular}{|c| l l l l l|}
            \hline\hline
            & \multicolumn{1}{c}{\multirow{3}{*}{\makecell{\textbf{Method}}}} & \multirow{3}{*}{\makecell{\textbf{Base} \\ \textbf{Acc.}}} &  \multirow{3}{*}{\makecell{\textbf{Prune} \\ \textbf{Acc.}}} & \multirow{3}{*}{\makecell{\textbf{Comp.} \\ \textbf{Rate}}} & \multirow{3}{*}{\makecell{\textbf{Sparsity (Method)} \\ \textbf{Scheme}}} \\
            &&&&& \multicolumn{1}{c|}{} \\
            &&&&& \multicolumn{1}{c|}{} \\\cline{1-6}
            % \multirow{6}{*}{\textbf{\rotatebox[origin=c]{90}{ResNet-18}}} & DCP~\cite{zhuang2018discrimination} & 88.9\% & 87.6\% & 2.0$\times$ & Structured \\
            \multirow{5}{*}{\textbf{\rotatebox[origin=c]{90}{ResNet-18}}} & AMC~\shortcite{he2018amc} & 90.5\% & 90.2\% & 2.0$\times$ &Channel (Lasso) \\
            % & Vari. Pruning~\shortcite{zhao2019variational} & 92.0\% & 91.7\% & $1.6\times$ & \textcolor{red}{XX} (Lasso)  \\
            & TinyADMM~\shortcite{ma2019tiny} & 94.1\% & 93.2\% & $15.1\times$ & Row+Col. (ADMM)  \\
            & \textbf{Our's} & \textbf{94.0\%} & \textbf{94.0\%} & \textbf{18.1$\times$} & One BLK (REW) \\
            % & \textbf{Our's} & \textbf{94.0\%} & \textbf{94.4\%} & \textbf{22.9$\times$} & $4\times 16$ BLK (REW) \\
            & \textbf{Our's} & \textbf{94.0\%} & \textbf{94.1\%} & \textbf{22.8$\times$} & $4\times 16$ BLK (REW) \\
            & \textbf{Our's} & \textbf{94.0\%} & \textbf{93.7\%} & \textbf{28.5$\times$} & $4\times 16$ BLK (REW) \\
            \hline
            \multirow{4}{*}{\textbf{\rotatebox[origin=c]{90}{MBNT}}} & DCP~\shortcite{zhuang2018discrimination} & 94.5\%  & 94.7\% & 1.4$\times$ & Channel (Heuristic) \\
            % & NISP~\cite{yu2018nisp} & \% & 93.2\% & 1.7$\times$ & Structured \\
            & \textbf{Our's} & \textbf{94.5\%} & \textbf{94.5\%} & \textbf{7.1$\times$} & One BLK (REW) \\
            % & \textbf{Our's} & \textbf{94.5\%} & \textbf{94.7\%} & \textbf{6.0$\times$} & $4\times 16$ BLK (REW) \\
            % & \textbf{Our's} & \textbf{94.5\%} & \textbf{94.5\%} & \textbf{7.2$\times$} & $4\times 16$ BLK (REW) \\
            & \textbf{Our's} & \textbf{94.5\%} & \textbf{94.5\%} & \textbf{8.9$\times$} & $4\times 16$ BLK (REW) \\
            & \textbf{Our's} & \textbf{94.5\%} & \textbf{93.4\%} & \textbf{10.3$\times$} & $4\times 16$ BLK (REW) \\
            \hline
            \multirow{5}{*}{\textbf{\rotatebox[origin=c]{90}{VGG-16}}} & 2PFPCE~\shortcite{min20182pfpce} & 92.9\% & 92.8\% & 4.0$\times$ & Row (Lasso) \\
            % & One Shot~\shortcite{liu2018rethinking} & 92.5\% & 92.4\% & 2.5$\times$ & Irregular \textcolor{red}{(XX)} \\
            % & 2PFPCE~\shortcite{min20182pfpce} & 92.9\% & 92.8\% & 4.0$\times$ & \textcolor{red}{XX} (Lasso) \\
            % & Effi. ConvNet~\shortcite{li2016pruning} & 93.2\% & 93.4\% & 2.7$\times$ & \textcolor{red}{XX} (Lasso) \\
            % & DCP~\cite{zhuang2018discrimination} & 93.9\% & 94.2\% & 2.0$\times$ & Structured \\
            % & FPGM~\cite{he2019filter} & 93.5\% & 93.2\% & 1.6$\times$ & Structured \\
            & TinyADMM~\shortcite{ma2019tiny} & 93.7\% & 92.7\% & $50.0\times$ & Row+Col. (ADMM)  \\
            & \textbf{Our's} & \textbf{93.5\%} & \textbf{93.0\%} & \textbf{50.0$\times$} & One BLK (REW) \\
            % & \textbf{Our's} & \textbf{93.5\%} & \textbf{93.8\%} & \textbf{35.7$\times$} & $4\times 16$ BLK (REW) \\
            & \textbf{Our's} & \textbf{93.5\%} & \textbf{93.5\%} & \textbf{50.1$\times$} & $4\times 16$ BLK (REW) \\
            & \textbf{Our's} & \textbf{93.5\%} & \textbf{93.0\%} & \textbf{69.7$\times$} & $4\times 16$ BLK (REW) \\
            \hline
					 
        \end{tabular}
    }
    \label{tab:cifar}
\end{table}

\begin{table}[t]
% \scriptsize
    \centering
    \caption{BLK-REW pruning results on ImageNet using VGG-16 and ResNet-18 and MobileNet-V2 (MBNT).}
    \resizebox{0.48 \textwidth}{!}{
        \begin{tabular}{|c| l l l l l|}
            \hline\hline
            & \multicolumn{1}{c}{\multirow{3}{*}{\makecell{\textbf{Method}}}} & \multirow{3}{*}{\makecell{\textbf{Base} \\ \textbf{Top-1/5} \\ \textbf{Acc.}}}&  \multirow{3}{*}{\makecell{\textbf{Prune} \\ \textbf{Top-1/5} \\ \textbf{Acc.}}} & \multirow{3}{*}{\makecell{\textbf{Comp.} \\ \textbf{Rate}}} & \multirow{3}{*}{\makecell{\textbf{Sparsity (Method)} \\ \textbf{Scheme}}} \\
            &&&&& \multicolumn{1}{c|}{} \\
            &&&&& \multicolumn{1}{c|}{} \\\cline{1-6}
            \multirow{7}{*}{\textbf{\rotatebox[origin=c]{90}{ResNet-18}}} & Network Slim.~\shortcite{liu2017learning} & 68.9/88.7\% & 67.2/87.4\% & 1.4$\times$ & Channel (Lasso) \\
            & DCP~\shortcite{zhuang2018discrimination} & 69.6/88.9\% & 64.1/85.7\% & 3.3$\times$ & Channel (Heuristic) \\
            & TinyADMM~\shortcite{ma2019tiny} & \emph{N/A}/89.1\% & \emph{N/A}/88.4\% & 3.3$\times$ & Row+Col. (ADMM) \\
            & StructADMM~\shortcite{zhang2018adam} & 69.9\%/\emph{N/A} & 68.8\%/\emph{N/A}\% & 3.0$\times$ & Col. (ADMM) \\
            & \textbf{Our's} & \textbf{69.9/89.1\%} & \textbf{69.0/88.5\%} & \textbf{4.0$\times$} & One BLK (REW) \\
            & \textbf{Our's} & \textbf{69.9/89.1\%} & \textbf{69.2/88.9\%} & \textbf{4.0$\times$} & $4\times 16$ BLK (REW) \\
            & \textbf{Our's} & \textbf{69.9/89.1\%} & \textbf{66.6/87.1\%} & \textbf{7.6$\times$} & $4\times 16$ BLK (REW) \\
            \hline
            \multirow{3}{*}{\textbf{\rotatebox[origin=c]{90}{MBNT}}} & AMC~\shortcite{he2018amc} & 71.8\%/\emph{N/A} & 70.8\%/\emph{N/A} & 1.4$\times$ &  Channel (Lasso) \\
            & \textbf{Our's} & \textbf{70.9/90.4\%} & \textbf{70.5/89.8\%} & \textbf{1.6$\times$} & One BLK (REW) \\
            & \textbf{Our's} & \textbf{70.9/90.4\%} & \textbf{70.0/89.7\%} & \textbf{2.0$\times$} & $4\times 16$ BLK (REW) \\
            \hline
            \multirow{6}{*}{\textbf{\rotatebox[origin=c]{90}{VGG-16}}} & Decorrelation~\shortcite{zhu2018ijcai} & 73.1\%/\emph{N/A} & 73.2\%/\emph{N/A} & 3.9$\times$ & Row (Group Lasso)\\
            & APoZ~\shortcite{hu2016network} & \emph{N/A}/88.4\% & 66.2/87.6\% & 2.0$\times$ & Channel (Heuristic) \\
            & AutoADMM~\shortcite{liu2019autoslim} & \emph{N/A}/92.1\% & \emph{N/A}/91.5\% & 6.4$\times$ & Row+Col. (ADMM) \\
            & \textbf{Our's} & \textbf{74.5/91.7\%} & \textbf{74.0/91.5\%} & \textbf{6.5$\times$} & One BLK (REW) \\
            & \textbf{Our's} & \textbf{74.5/91.7\%} & \textbf{74.4/91.6\%} & \textbf{3.1$\times$} & $4\times 16$ BLK (REW) \\
            & \textbf{Our's} & \textbf{74.5/91.7\%} & \textbf{73.8/91.2\%} & \textbf{7.8$\times$} & $4\times 16$ BLK (REW) \\
            \hline
					 
        \end{tabular}
    }
    \label{tab:imagenet}
\end{table}

\begin{table}[!t]
\caption{BLK-REW pruning results comparison on GRU with TIMIT dataset. PER denotes {\em phone error rate.}}
\resizebox{0.48 \textwidth}{!}{
\label{tab:timit}
\begin{tabular}{|l l l l l l|}
\hline
\multicolumn{1}{|c}{\multirow{3}{*}{\makecell{\textbf{Method}}}} & \multirow{3}{*}{\makecell{\textbf{Base} \\ \textbf{PER}}} &  \multirow{3}{*}{\makecell{\textbf{Prune} \\ \textbf{PER}}} & \multirow{3}{*}{\makecell{\textbf{Comp.} \\ \textbf{Rate}}} & \multirow{3}{*}{\makecell{\textbf{Sparsity} \\ \textbf{(Method)} \\ \textbf{Scheme}}} & \multirow{3}{*}{\makecell{\textbf{Exe.} \\ \textbf{Time ($ms$)} \\ \textbf{CPU/GPU}}} \\
&&&&& \multicolumn{1}{c|}{} \\
&&&&& \multicolumn{1}{c|}{} \\\cline{1-6}
ESE~\shortcite{han2017ese} & 20.40\% & 20.70\% & 8.0$\times$ & Irregular (Heuristic) & \emph{N/A} \\
% C-LSTM~\shortcite{wang2018c} & 24.15\% & 24.57\% & 8.0$\times$ & Block-circulant & \emph{N/A} \\
C-LSTM~\shortcite{wang2018c} & 24.15\% &  25.48\% & 16.0$\times$ & Block-circ. & \emph{N/A} \\
E-RNN~\shortcite{li2019ERNN} & 20.02\% & 20.20\% & 8.0$\times$ & Block-circ. & \emph{N/A} \\
%  & Pconv & 93.5\% &  &  & BCR (4 $\times$ 16) \\
%  \textbf{BCR Pruning} & 93.5\% & 94.5\% & 18.0$\times$ & BCR (4 $\times$ 16) \\
\textbf{Our's} & \textbf{18.8\%} & \textbf{18.8\%} & \textbf{19.1$\times$} & BLK (REW) & \textbf{0.97/0.50} \\ 
\textbf{Our's} & \textbf{18.8\%} & \textbf{23.2\%} & \textbf{112.9$\times$} & BLK (REW) & \textbf{0.35/0.25} \\ 
\textbf{Our's} & \textbf{18.8\%} & \textbf{24.0\%} & \textbf{231.3$\times$} & BLK (REW) & \textbf{0.21/0.09} \\
\hline
\end{tabular}
}\vspace{-3.0ex}
\end{table}

\subsection{Performance Evaluation on Mobile Devices}

\textbf{Execution time} results are shown in Figure~\ref{fig:speed}. We test the BLK pruned model on mobile CPU/GPU. To ensure fairness, all frameworks are using the same pattern-based sparse model, and we also enable the fully optimized configurations of TFLite, TVM and MNN (e.g., Winograd optimization is turned on). All test models are the ones with the largest compression rates in Table~\ref{tab:imagenet} and Table~\ref{tab:cifar}. For GRU RNN execution, since other frameworks do not support end-to-end execution on mobile devices, we only report the execution time of the proposed block-based sparse model with block size $2\times 32$ in Table~\ref{tab:timit}. 
We can see our approach achieves significant acceleration on mobile devices compared with other frameworks. For image classification tasks, all of our results on mobile GPU exceed the real-time requirements (i.e., usually 33$ms$/frame). For NLP tasks, the proposed framework also achieves real-time speech recognition.

% \section{Discussion on Generality}

\section{Conclusion}
This paper presents the block-based DNN structured pruning framework using reweighted regularization method (BLK-REW). The proposed block-based structured sparsity is flexible and can be used in both CNN and RNN applications. With the support of the compiler code generation and optimization, our framework can achieve real-time acceleration on many devices. 
The proposed framework also uses reweighted method to dynamically update the regularization process, which improves the pruning results effectively within considerably shorter training time. Compared with state-of-the-art pruning methods, the proposed framework is general and achieves high performance.

{
% \bibliographystyle{named}
% \bibliography{ijcai20}

\begin{thebibliography}{}

\bibitem[\protect\citeauthoryear{Ali}{}]{Ali-MNN}
\url{https://github.com/alibaba/MNN}.

\bibitem[\protect\citeauthoryear{Boyd \bgroup \em et al.\egroup
  }{2011}]{boyd2011distributed}
Stephen Boyd, Neal Parikh, Eric Chu, Borja Peleato, and Jonathan Eckstein.
\newblock Distributed optimization and statistical learning via the alternating
  direction method of multipliers.
\newblock {\em Foundations and Trends{\textregistered} in Machine Learning},
  2011.

\bibitem[\protect\citeauthoryear{Candes \bgroup \em et al.\egroup
  }{2008}]{candes2008enhancing}
Emmanuel~J Candes, Michael~B Wakin, and Stephen~P Boyd.
\newblock Enhancing sparsity by reweighted l1 minimization.
\newblock {\em Journal of Fourier analysis and applications}, 2008.

\bibitem[\protect\citeauthoryear{Chen \bgroup \em et al.\egroup
  }{2018}]{chen2018tvm}
Tianqi Chen, Thierry Moreau, Ziheng Jiang, Lianmin Zheng, Eddie Yan, Haichen
  Shen, Meghan Cowan, Leyuan Wang, Yuwei Hu, Luis Ceze, et~al.
\newblock {TVM}: An automated end-to-end optimizing compiler for deep learning.
\newblock In {\em OSDI}, 2018.

\bibitem[\protect\citeauthoryear{Gehring \bgroup \em et al.\egroup
  }{2016}]{gehring2016convolutional}
Jonas Gehring, Michael Auli, David Grangier, and Yann~N Dauphin.
\newblock A convolutional encoder model for neural machine translation.
\newblock {\em arXiv preprint arXiv:1611.02344}, 2016.

\bibitem[\protect\citeauthoryear{Han \bgroup \em et al.\egroup
  }{2017}]{han2017ese}
Song Han, Junlong Kang, Huizi Mao, Yiming Hu, Xin Li, Yubin Li, Dongliang Xie,
  Hong Luo, Song Yao, Yu~Wang, Huazhong Yang, and William~J. Dally.
\newblock Ese: Efficient speech recognition engine with sparse lstm on fpga.
\newblock In {\em FPGA}, 2017.

\bibitem[\protect\citeauthoryear{He \bgroup \em et al.\egroup
  }{2016}]{he2016deep}
Kaiming He, Xiangyu Zhang, Shaoqing Ren, and Jian Sun.
\newblock Deep residual learning for image recognition.
\newblock In {\em CVPR}, 2016.

\bibitem[\protect\citeauthoryear{He \bgroup \em et al.\egroup
  }{2017}]{he2017channel}
Yihui He, Xiangyu Zhang, and Jian Sun.
\newblock Channel pruning for accelerating very deep neural networks.
\newblock In {\em ICCV}, 2017.

\bibitem[\protect\citeauthoryear{He \bgroup \em et al.\egroup
  }{2018}]{he2018amc}
Yihui He, Ji~Lin, Zhijian Liu, Hanrui Wang, Li-Jia Li, and Song Han.
\newblock Amc: Automl for model compression and acceleration on mobile devices.
\newblock In {\em ECCV}, 2018.

\bibitem[\protect\citeauthoryear{Howard \bgroup \em et al.\egroup
  }{2017}]{howard2017mobilenets}
Andrew Howard, Menglong Zhu, Bo~Chen, Dmitry Kalenichenko, Weijun Wang, Tobias
  Weyand, Marco Andreetto, and Hartwig Adam.
\newblock Mobilenets: Efficient convolutional neural networks for mobile vision
  applications.
\newblock {\em arXiv:1704.04861}, 2017.

\bibitem[\protect\citeauthoryear{Hu \bgroup \em et al.\egroup
  }{2016}]{hu2016network}
Hengyuan Hu, Rui Peng, Yu-Wing Tai, and Chi-Keung Tang.
\newblock Network trimming: A data-driven neuron pruning approach towards
  efficient deep architectures.
\newblock {\em arXiv preprint arXiv:1607.03250}, 2016.

\bibitem[\protect\citeauthoryear{Li \bgroup \em et al.\egroup
  }{2019}]{li2019ERNN}
Zhe Li, Caiwen Ding, Shuo Wang, Wujie Wen, Youwei Zhuo, Xue Lin, Xuehai Qian,
  and Yanzhi Wang.
\newblock E-rnn: design optimization for efficient recurrent neural networks in
  fpgas.
\newblock In {\em HPCA}, 2019.

\bibitem[\protect\citeauthoryear{Liu \bgroup \em et al.\egroup
  }{2017}]{liu2017learning}
Zhuang Liu, Jianguo Li, Zhiqiang Shen, Gao Huang, Shoumeng Yan, and Changshui
  Zhang.
\newblock Learning efficient convolutional networks through network slimming.
\newblock In {\em ICCV}, 2017.

\bibitem[\protect\citeauthoryear{Liu \bgroup \em et al.\egroup
  }{2020}]{liu2019autoslim}
Ning Liu, Xiaolong Ma, Zhiyuan Xu, Yanzhi Wang, Jian Tang, and Jieping Ye.
\newblock Autoslim: An automatic dnn structured pruning framework for
  ultra-high compression rates.
\newblock {\em AAAI}, 2020.

\bibitem[\protect\citeauthoryear{Ma \bgroup \em et al.\egroup
  }{2020}]{ma2019tiny}
Xiaolong Ma, Geng Yuan, Sheng Lin, Caiwen Ding, Fuxun Yu, Tao Liu, Wujie Wen,
  Xiang Chen, and Yanzhi Wang.
\newblock Tiny but accurate: A pruned, quantized and optimized memristor
  crossbar framework for ultra efficient dnn implementation.
\newblock {\em ASP-DAC}, 2020.

\bibitem[\protect\citeauthoryear{Min \bgroup \em et al.\egroup
  }{2018}]{min20182pfpce}
Chuhan Min, Aosen Wang, Yiran Chen, Wenyao Xu, and Xin Chen.
\newblock 2pfpce: Two-phase filter pruning based on conditional entropy.
\newblock {\em arXiv:1809.02220}, 2018.

\bibitem[\protect\citeauthoryear{Nugraha \bgroup \em et al.\egroup
  }{2017}]{nugraha2017towards}
Brilian~Tafjira Nugraha, Shun-Feng Su, et~al.
\newblock Towards self-driving car using convolutional neural network and road
  lane detector.
\newblock In {\em ICACOMIT}, 2017.

\bibitem[\protect\citeauthoryear{Paszke \bgroup \em et al.\egroup
  }{2019}]{paszke2019pytorch}
Adam Paszke, Sam Gross, Francisco Massa, Adam Lerer, James Bradbury, Gregory
  Chanan, Trevor Killeen, Zeming Lin, Natalia Gimelshein, Luca Antiga, et~al.
\newblock Pytorch: An imperative style, high-performance deep learning library.
\newblock In {\em NeurIPS}, 2019.

\bibitem[\protect\citeauthoryear{Ren \bgroup \em et al.\egroup
  }{2019}]{ren2019ADMMNN}
Ao~Ren, Tianyun Zhang, Shaokai Ye, Wenyao Xu, Xuehai Qian, Xue Lin, and Yanzhi
  Wang.
\newblock Admm-nn: an algorithm-hardware co-design framework of dnns using
  alternating direction methods of multipliers.
\newblock In {\em ASPLOS}, 2019.

\bibitem[\protect\citeauthoryear{Simonyan and
  Zisserman}{2014}]{simonyan2014very}
Karen Simonyan and Andrew Zisserman.
\newblock Very deep convolutional networks for large-scale image recognition.
\newblock {\em arXiv:1409.1556}, 2014.

\bibitem[\protect\citeauthoryear{Ten}{}]{TensorFlow-Lite}
\url{https://www.tensorflow.org/mobile/tflite/}.

\bibitem[\protect\citeauthoryear{Wang \bgroup \em et al.\egroup
  }{2018}]{wang2018c}
Shuo Wang, Zhe Li, Caiwen Ding, Bo~Yuan, Qinru Qiu, Yanzhi Wang, and Yun Liang.
\newblock C-lstm: Enabling efficient lstm using structured compression
  techniques on fpgas.
\newblock In {\em FPGA}, 2018.

\bibitem[\protect\citeauthoryear{Wen \bgroup \em et al.\egroup
  }{2016}]{wen2016learning}
Wei Wen, Chunpeng Wu, Yandan Wang, Yiran Chen, and Hai Li.
\newblock Learning structured sparsity in deep neural networks.
\newblock In {\em NeurIPS}, 2016.

\bibitem[\protect\citeauthoryear{Yuan and Lin}{2006}]{yuan2006model}
Ming Yuan and Yi~Lin.
\newblock Model selection and estimation in regression with grouped variables.
\newblock {\em Statistical Methodology}, 2006.

\bibitem[\protect\citeauthoryear{Zhang \bgroup \em et al.\egroup
  }{2018a}]{zhang2018systematic}
Tianyun Zhang, Shaokai Ye, Kaiqi Zhang, Jian Tang, Wujie Wen, Makan Fardad, and
  Yanzhi Wang.
\newblock A systematic dnn weight pruning framework using alternating direction
  method of multipliers.
\newblock In {\em ECCV}, 2018.

\bibitem[\protect\citeauthoryear{Zhang \bgroup \em et al.\egroup
  }{2018b}]{zhang2018adam}
Tianyun Zhang, Kaiqi Zhang, Shaokai Ye, Jian Tang, Wujie Wen, Xue Lin, Makan
  Fardad, and Yanzhi Wang.
\newblock Adam-admm: A unified, systematic framework of structured weight
  pruning for dnns.
\newblock {\em arXiv preprint arXiv:1807.11091}, 2:3, 2018.

\bibitem[\protect\citeauthoryear{Zhu \bgroup \em et al.\egroup
  }{2018}]{zhu2018ijcai}
Xiaotian Zhu, Wengang Zhou, and Houqiang Li.
\newblock Improving deep neural network sparsity through decorrelation
  regularization.
\newblock In {\em IJCAI}, 2018.

\bibitem[\protect\citeauthoryear{Zhuang \bgroup \em et al.\egroup
  }{2018}]{zhuang2018discrimination}
Zhuangwei Zhuang, Mingkui Tan, Bohan Zhuang, Jing Liu, Yong Guo, Qingyao Wu,
  Junzhou Huang, and Jinhui Zhu.
\newblock Discrimination-aware channel pruning for deep neural networks.
\newblock In {\em NeurIPS}, 2018.

\end{thebibliography}

}

\end{document}